\documentclass[10pt,twocolumn,letterpaper]{article}

\usepackage{cvpr}
\usepackage{times}
\usepackage{epsfig}
\usepackage{graphicx}
\usepackage{subfigure}
\usepackage{amsmath}
\usepackage{amssymb}
\usepackage{amsthm}
\usepackage{amsfonts}
\usepackage{mathrsfs}
\usepackage{booktabs}
\usepackage{multirow, eucal}
\usepackage{slashbox,pict2e}

\usepackage[vlined,ruled,linesnumbered]{algorithm2e}

\usepackage{cite}

\usepackage[pagebackref=true,breaklinks=true,letterpaper=true,colorlinks,bookmarks=false]{hyperref}

\usepackage{caption}
\newcommand*\samethanks[1][\value{footnote}]{\footnotemark[#1]}

\cvprfinalcopy %

\ifcvprfinal\fi

\begin{document}
\title{
Attend in groups: a weakly-supervised deep learning framework for learning from web data}
\author{Bohan Zhuang\thanks{First two authors contributed  equally. Correspondence should be addressed to C. Shen.}, ~ Lingqiao Liu\samethanks, ~ Yao Li, ~ Chunhua Shen, ~ Ian Reid\\
	The University of Adelaide; and Australian Centre for Robotic Vision
}
\maketitle
\thispagestyle{empty}
\begin{abstract}

Large-scale datasets have driven the rapid development of deep neural networks for visual recognition. However, annotating a massive dataset is expensive and time-consuming. Web images and their labels are, in comparison, much easier to obtain, but direct training on such automatically harvested images can lead to unsatisfactory performance, because the noisy labels of Web images adversely affect the learned recognition models. To address this drawback we propose an end-to-end weakly-supervised deep learning framework which is robust to the label noise in Web images. The proposed framework relies on two unified strategies -- random grouping and attention -- to effectively reduce the negative impact of noisy web image annotations.  Specifically, random grouping stacks multiple images into a single training instance and thus increases the labeling accuracy at the instance level. Attention, on the other hand, suppresses the noisy signals from both incorrectly labeled images and less discriminative image regions.  By conducting intensive experiments on two challenging datasets, including a newly collected fine-grained dataset with Web images of different car models, the superior performance of the proposed methods over competitive baselines is clearly demonstrated.

\end{abstract}

\tableofcontents
\clearpage

\section{Introduction}

Recent development of deep convolutional neural networks (CNNs) has led to great success in a variety of tasks including image classfication~\cite{krizhevsky2012imagenet, simonyan2014very, he2015deep}, object detection~\cite{girshick2014rich, ren2015faster, liu2015ssd}, semantic segmentation~\cite{long2015fully,  Lin_2016_CVPR} and others. This success is largely driven by the availability of large-scale well-annotated image datasets, e.g. ImageNet~\cite{ILSVRC15}, MS COCO~\cite{lin2014microsoft} and PASCAL VOC~\cite{everingham2010pascal}. However, annotating a massive number of images is extremely labor-intensive and costly. To reduce the annotating labor cost, an alternative approach is to obtain the image annotations directly from the image search engine from the Internet,  e.g. Google image search or Bing images.

Web-scale image search engine mostly uses keywords as queries and the connection between keywords and images is established by the co-occurrence between the Web image and its surrounding text. Thus, the annotations of Web images returned by a search engine will be inevitably noisy since the query keywords may not be consistent with the visual content of target images. For example, using ``black swan'' as a query keyword, the retrieved images may contain ``white swan,'' ``swan painting'' and some other different categories. These noisy labels can be misleading if we use them to train a classifier to learn the corresponding visual concept.

To overcome this drawback, we propose a deep learning framework designed to be more robust to the labeling noise and thus better able to leverage Web images for training. There are two key strategies in our framework: random grouping and attention. As will be shown later, these two strategies seamlessly work together to reduce the negative impact of label noise.

Specifically, the random grouping strategy randomly samples a few images and merges them into a single training instance. The idea is that although the probability of sampling an incorrectly labeled Web image is high, the probability of sampling an incorrectly labeled group is low because as long as one image in the group is correctly labeled, the label of the group is deemed correct (bag label as in multi-instance learning). In the proposed approach, each image is represented by the extracted contextual features depicting the visual patterns of local image regions. After the random grouping, a training instance is represented as the union of convolutional feature maps extracted from each image in the group. If there are any incorrectly labeled images in the group, the unified feature maps of an instance will contain a substantial amount of local features which are irrelevant to the group-level class annotation. To avoid the distraction of those local features, we apply the second strategy of our framework, the attention mechanism, to encourage the network not to focus on the irrelevant features.

\begin{figure*}[tbp]
		\centering
		\resizebox{\linewidth}{!}
		{\includegraphics{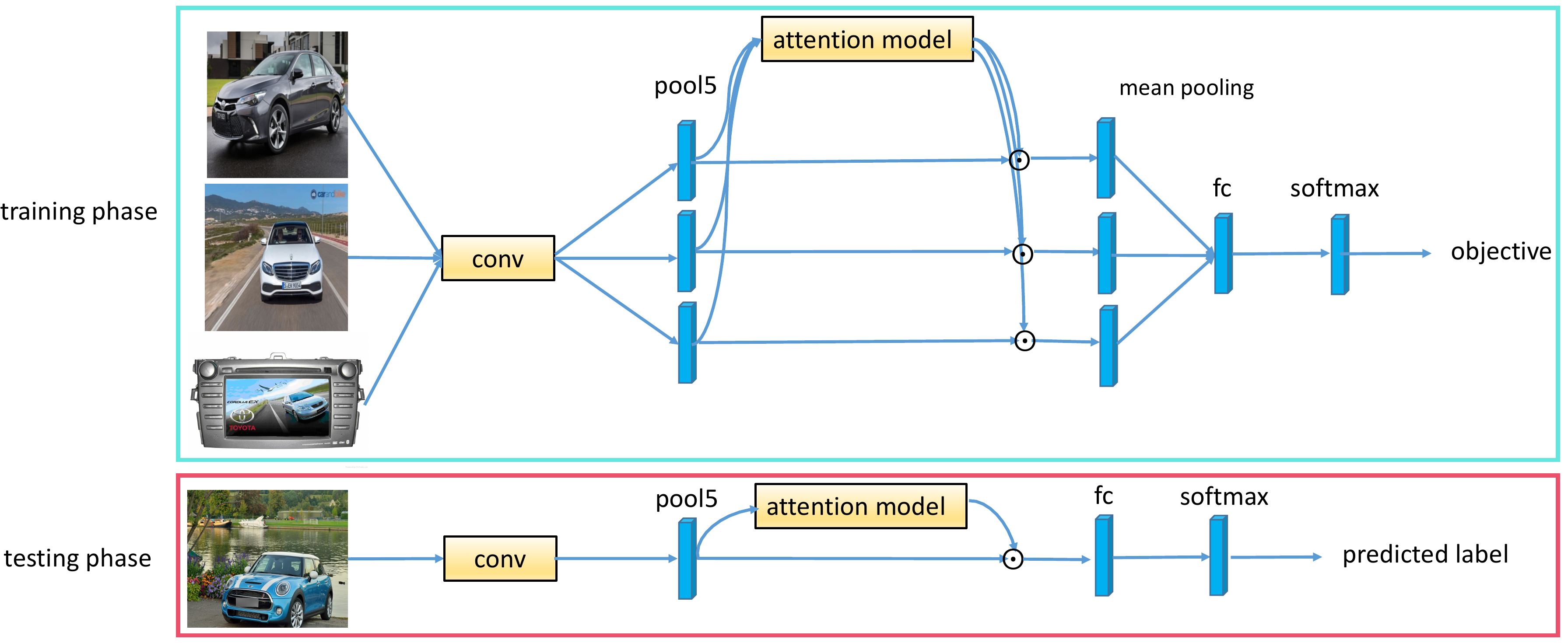}}
		\caption{ Overview of our ``webly''-supervised learning pipeline. For the training phase, inputs are a group of images, including one correctly labeled image and two noise images from top to bottom. The convolutional layers are shared. The attention model is added on each training data and followed by a global average pooling layer to get the aggregated group-level representation, followed by a softmax layer for classification. For the testing phase, the input is a single image and output is the predicted class label.
		}
		\label{fig:overview}
\end{figure*}

To experimentally validate the robustness of the proposed method, we collect a large-scale car dataset using a Web image search engine. This dataset is particularly challenging due to its fine-grained nature. By conducting an experimental comparison on this dataset, we demonstrate that the proposed method achieves significantly better performance than competitive approaches.

\section{Related Work}

Our work is closely related to learning from web-scale data and noisy labels~\cite{fergus2010learning, schroff2011harvesting, xu2015augmenting, chen2015webly, divvala2014learning, krause2015unreasonable, chen2013neil, niu2015visual, reed2014training,  sukhbaatar2014learning, xiao2015learning, mnih2012learning}.  In terms of learning from Web data, in~\cite{chen2013neil, chen2015webly}, Chen \etal propose to pretrain CNN on simple examples and adapt it to harder images by leveraging the structure of data and categories in a two-step manner. In contrast, we propose a simply-yet-effective end-to-end learning framework without pretraining. To better dealing with noise, some approaches~\cite{xiao2015learning, sukhbaatar2014training} propose to add an extra noise layer into the network which adapts the network outputs to match the noisy label distribution. On the other hand, some approaches attempt to remove or correct noisy labels~\cite{brodley2011identifying, miranda2009use}, but because of the difficulty of separating correctly labeled hard samples from mislabeled ones, such a strategy can result in removing too many (correct) instances. Moreover, several label noise-robust algorithms~\cite{beigman2009learning, manwani2013noise} are proposed to make classifiers robust to label noise. However, noise-robust methods seem to be adequate only for simple cases of label noise that can be safely managed by regularization. In this paper, we instead propose to suppress label noise by unified two strategies without any strong assumptions.

Our work is also related to weakly-supervised object localization~\cite{oquab2015object, rochan2015weakly, deselaers2012weakly, song2014learning, wang2014weakly, ren2016weakly, bilen2015weakly, cinbis2014multi}. The objective of these methods is to localize object parts that are visually consistent with the semantic image-level labels across the training data. A major approach for tackling this task is to formulate it as a multiple instance learning problem. In these methods~\cite{cinbis2014multi, ren2016weakly, wu2015deep}, each image is modeled as a bag of instances (region features) and the classifier is learned to select the foreground instances. Further in \cite{oquab2015object}, a weakly-supervised deep learning pipeline is proposed to localize objects from complex cluttered scenes by explictly searching over possible object locations and scales in the image. In light of the above methods, we convert the problem of learning from noisy labels to a weakly-supervised problem that in spirit is similar to the multiple instance learning assumption. What's more, we further propose to incorporate attention strategy to reduce the adverse effect of noise.

Related to our work, the attentive mechanisms have been applied to many computer vision tasks~\cite{ba2014multiple, mnih2014recurrent, xu2015show, yang2015stacked, andreas2015deep, xu2015ask, bahdanau2014neural, weston2014memory, graves2014neural, lu2016hierarchical} to help improve the performance. To guide the models' focus on the objects specified by the question or caption, attention models are designed to pay attention to local CNN features in the input image~\cite{xu2015show,yang2015stacked, andreas2015deep, xu2015ask, lu2016hierarchical}. The attentive mechanism has also been used to handle sequential problems in neural machine translation~\cite{bahdanau2014neural, luong2015effective} and manage memory access mechanisms for memory networks~\cite{weston2014memory} and neural turing machines~\cite{graves2014neural}. Different from the above methods, we are the first to apply the attention mechanism to cope with noisy labels. It can not only detect discriminative local feature regions, but also serves to filter out noisy signals from the mislabeled samples in the training instance.

\section{Method}

In our task, we intend to distill useful visual knowledge from the noisy Web data. It consists of correctly labeled samples and mislabeled samples on the Web. To make the classifier robust to noisy labels, we propose a deep learning framework by incorporating two strategies, random group training, and attention. The overview of our method is shown in Figure \ref{fig:overview}. At the training stage, we randomly group multiple training images into a single training instance as the input of our neural network. The proposed neural network architecture has two parts. The first part is similar to a standard convolutional neural network which is comprised of multiple convolutional layers and pooling layers. The second part is an attentional pooling layer which selects parts of the neuron activations and pools the activations into the instance-level representation. Once the neural network is trained, we can drop off the random grouping module and takes a single image as input at the test stage.

In the following sections, we will elaborate the random grouping training and the attention module and discuss their benefits for reducing the impact of noisy labels.

\subsection{Random grouping training} \label{sec:weakly}

Random grouping training (RGT) aims at reducing the probability of sampling an incorrectly labeled instance and thus mitigate the risk confusing a neural work with wrong annotations. The idea of RGT is to stack multiple images of one class into a single grouped training instance of the same class. In practice, we implement this idea by stacking the last layer convolutional feature maps obtained from each image into a unified convolutional feature map and perform (attention based) pooling on this feature map to obtain the instance-level  representation. In this sense, we can view the input of a grouped instance as a ``merged image'' and as long as one image is correctly labeled as containing the object-of-interest, the ``merged image'' indeed contains it.  In other words, the grouped training instance is correctly labeled as long as one image within is correctly labeled.

Consequently, if the probability of sampling an incorrectly labeled image is $\xi $, then the probability of sampling a correctly labeled grouped instance will become
\begin{align}
p = 1 - \xi ^K
\end{align}where $K$ is the group size and when $K$ becomes larger, the probability of sampling a correctly labeled instance will become very high. For example, if $\xi=0.2$ and $K=3$, $p$ will be greater than 99\%. However, when $K$ becomes larger, the independence between multiple training instances will reduce and this tends to undermine the network training. Thus in practice, we choose $K$ as a small value (2 to 5). We have conducted an experimental study on the impact of $K$ with respect to different level of labeling noise at Section \ref{sec:group_size}.

\begin{figure*}[tbp]
	\centering
	\resizebox{1.0\linewidth}{!}
	{\begin{tabular}{c}
			\includegraphics[width=0.50\linewidth, height=0.25\linewidth]{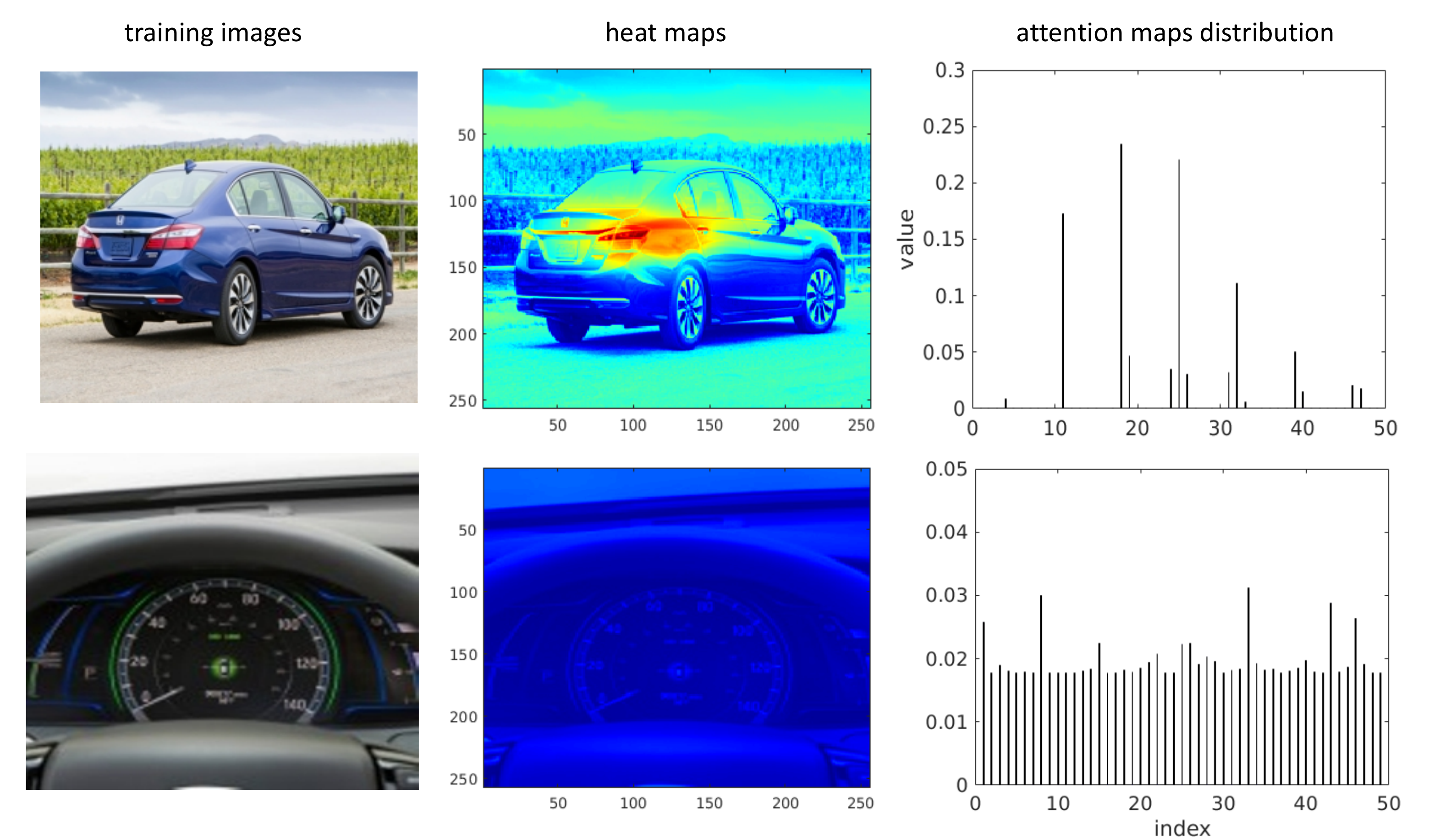}
		\end{tabular}
	}
	\caption{This figure illustrates the effectiveness of the group-wise attention model used in the proposed method. The left column shows the original training images. The middle column is the images plus its corresponding attention heat maps. The right column shows the distribution of the attention maps.  The upper row relates to the correctly labeled sample and the bottom row corresponds to the mislabeled sample. We can see that for the correctly labeled sample, the normalized attention model only focus on the discriminative local parts and the score distribution is sparse. In contract, for the mislabeled sample, the normalized attention model fails to concentrate on any local regions and the score distribution is dense. }
	\label{fig:attention}
\end{figure*}

\subsection{Attention}
\subsubsection{Attention formulation}
After random grouping, each instance is now represented as an array of activations. These activations come from both correctedly labeled images and mislabeled images.  Although containing activations from the correct region of interest, many of the activations are noisy signals and will negatively impact the learning process. To mitigate this issue, we propose to use an attention model to focus processing only on the attended activations.
Let ${\bf{x}}_{ijk}^n \in {\mathbb{R}^{c}}$ denote the last convolutional layer activations from the $k$-th image of the $n$-th instance at the spatial location $(i,j)$, where $i=1,2,...,d$ and $j=1,2,...,d$ are the coordinates of the feature map and $d$ is the height or width of the feature map. %

The unnormalized attention score ${s}_{ijk}^n \in {\mathbb{R}}$ can be formulated as
\begin{equation} \label{eq:detector}
{s}_{ijk}^n = f({{\bf{w}}^{\rm{T}}}{\bf{x}}_{ijk}^n + b),
\end{equation}
where ${\bf{w}} \in {\mathbb{R}^c}$, $b \in {\mathbb{R}^1}$ denote the weight and bias of the attention detector respectively, which are parts of the model parameters and will be learned in an end-to-end manner. $f(\cdot)$ is the softplus function $f(x) = \ln (1 + \exp (x))$. %
Since we are only concerned with the relative importance of the local features within an image, we propose to normalize the attention scores to $[0,1]$ for aggregating the local features:
\begin{equation}
{a}_{ijk}^n = \frac{{{s}_{ijk}^n + \varepsilon }}{{\sum\limits_i {\sum\limits_j {({s}_{ijk}^n + \varepsilon )} } }},
\end{equation}
where ${a}_{ijk}^n$ is the normalized attention score, $\varepsilon$ is a small constant and quite important to make the distribution reasonable.

If the element ${s}_{ijk}^n$ is low but there is no $\varepsilon$, then the corresponding ${a}_{ijk}^n$ can be large even though ${s}_{ijk}^n$ is small. The constant $\varepsilon$ can solve this problem effectively. If it is properly set, a small ${s}_{ijk}^n$ (approaching zero) will result in ${{a}_{ijk}^n} = \frac{1}{{{d^2}}}$. In our work, we set it to $0.1$.

After obtaining the normalized attention scores, we can get the attended feature representation by applying ${a}_{ijk}^n$ to ${\bf{x}}_{ijk}^n$ as follows:
\begin{equation}
	\widehat {\bf{x}}_{ijk}^n = {a}_{ijk}^n \odot {\bf{x}}_{ijk}^n,
\end{equation}
where $\odot$ is the element-wise multiplication, $\widehat {\bf{x}}_{ijk}^n$ is the attended feature representation.

Then the representation of a grouped training instance can be obtained by a global average pooling over all the feature dimensions except for the channel-wise dimension:
\begin{equation} \label{eq:sum_pool}
{{\bf{h}}_n} = \frac{1}{{{d^2}k}}\sum\limits_i {\sum\limits_j {\sum\limits_k {\widehat {\bf{x}}_{ijk}^n} } },
\end{equation}
where ${{\bf{h}}_n} \in {\mathbb{R}^c}$ is the group-level representation of the $n$-th training instance.

Then we apply a linear classifier layer to predict the class label of each grouped instance and use the multi-class cross-entropy loss to train the network:
\begin{equation}
	L_{class} =  - \sum\limits_n {{y_n}} \log (\frac{{\exp ({{\bf{F}}_n})}}{{\sum\limits_n {\exp ({{\bf{F}}_n})} }})
\end{equation}
where ${\bf{F}}_n$ and $y_n$ are the last linear classification layer and the class label for the $n$-th training instance, respectively.

\subsubsection{Attention module regularization } \label{sec:robust}
Ideally, for the correctly labeled image, the attention scores should have large values on one or few image regions; for the mislabeled image, none of the image regions should correspond to large attention values. In the above framework, we expect this situation can happen after the end-to-end training of the network. In this section, we devise a regularization term to further encourage this property. To apply this regularization, we assume that a set of negative class images belonging to none of to-be-learned image categories is available. Then we can apply the attention detector on those negative class images and require that the obtained normalized attention values are as small as possible since those images do not contain the object-of-interest. Define $u_{ijk}^n = {{\bf{w}}^{\rm{T}}}{\bf{x}}_{ijk}^n + b$ to be the linear attention scores for the sample ${\bf{x}}_{ijk}^n$ ; then the above requirement is equivalent to expecting $\max_{ijk}{u_{ijk}^n} < 0$. On the other hand, for a grouped training instance generated from each class, we expect that the attention detector identifies at least one relevant region and this leads to the objective $\max_{ijk}{u_{ijk}^n} > 0$. In this paper, we propose to use the following objective function to impose the aforementioned two requirements:
\begin{align}
	R({\bf{w}},{b}) = \sum\limits_n {\max (0,\,\,1 - {\delta _n}{{\max }_{ijk}}(u_{ijk}^n)} )
\end{align}where $\delta_n=\{1,-1\}$ indicates whether the instance is sampled from the classes of object-of-interest or from the negative class.
We then use the weighted sum of $L_{class}$ and $R$ as the final objective function:
\begin{align}
	L = {L_{class}} + \lambda R.
\end{align}

The effect of the attention module is illustrated in Figure~\ref{fig:attention}. The input is an instance including a correctly labeled car sample and a mislabeled noise sample. We can observe that for the correctly labeled sample, the normalized attention scores are pushed high at the region-of-interest, which corresponds to the back of the car in the example. In contrast, for the mislabeled sample, the normalized attention scores are all pushed approaching zero, resulting in no parts to be concentrated on for the attention model. In terms of this observation, we can explore that the attention model can not only filter out the contextual features of the mislabeled samples in the training instance, but also help detect the discriminative parts of the correctly labeled samples.

\section{Experiments}
In this section, we test our weakly-supervised learning framework on two datasets collected from the Web. One is a fine-grained dataset and the other one is a conventional classification dataset. The training data for both tasks are obtained via search results freely available from Google image search, using all returned images as training data.
It's worth noticing that fine-grained classification is quite challenging because categories can only be discriminated by subtle and local differences.

\subsection{Datasets} \label{sec:datasets}
\noindent \textbf{WebCars:}  We collect a large-scale fine-grained car dataset from the internet, named WebCars\footnote{The dataset will be available soon.}, using the categories of the clean CompCars dataset~\cite{yang2015large}. We treat the car model names as the query keywords and automatically retrieve images for all the 431 fine-grained categories. We collect 213,072 noisy Web images in total and still use the test set of the original clean dataset for testing. We sample a few categories from WebCars and mannually annotate the ground-truth labels, noting in the process that approximately $30\%$ of images are outliers. We further collect 10,000 images that doesn't belong to the training categories as the negative class.

\noindent \textbf{Web data + ImageNet:}  We randomly sample 100 classes used in ImageNet and use the category names for collecting a noisy Web image dataset. All the images are automatically downloaded and the ones that appear in the original ImageNet dataset are manually removed. This dataset contains 61,639 images in total. The noise gradually increases from the highly ranked images to the latter samples. We estimate the percentage of mislabeled samples is approximately 20 \%. We also collect 5,000 negative class Web images.

\begin{figure*}[tbp]
	\centering
	\resizebox{\linewidth}{!}
	{\includegraphics[width=1\linewidth, height=0.3\linewidth]{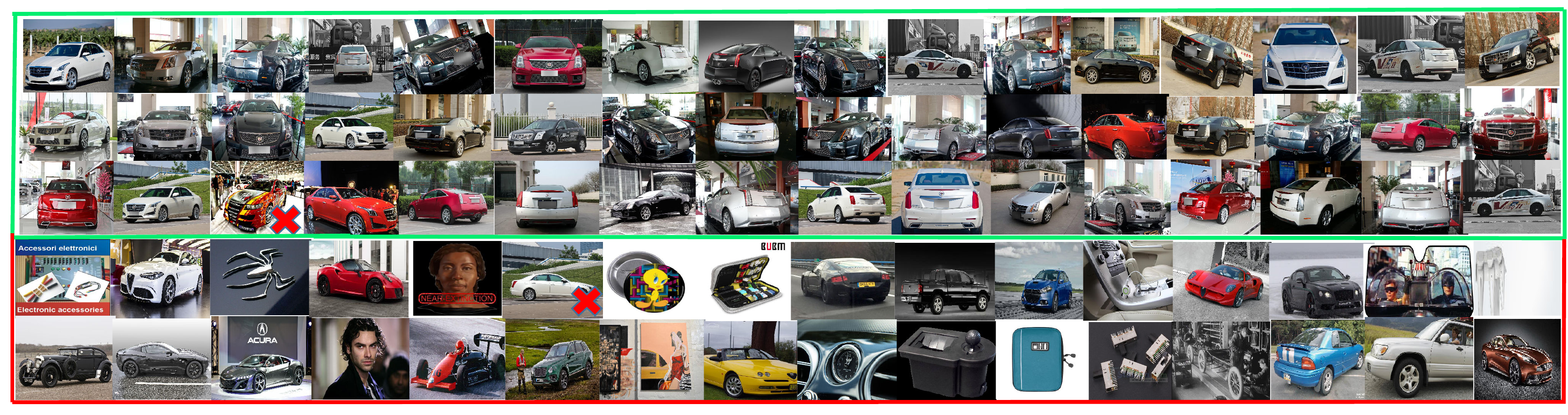}}
	\caption{Examples of the image re-ranking performance on one sampled car category (``cadillac''). The red crosses indicate the images that are classified incorrectly. The images are sorted according to the rank of the classification scores in descending order. The images in the green rectangle and red rectangle are correctly labeled samples and mislabeled samples,  respectively. The noise level is 0.4.}
	\label{fig:rerank}
\end{figure*}

\subsection{Implementation details} \label{sec:implementation}
We use Theano~\cite{bastien2012theano} for our experiments. We use the pretrained VGG-16 model trained on the ImageNet dataset \cite{ILSVRC15} to initialize the convolutional layers of our framework. The learning rate is set to 0.001 initially, and divided by 10 after 5 epoches.  The regularizer $\lambda$ is set to 0.1. Training samples are randomly grouped online.

To investigate the impact of the various elements in our end-to-end framework, we analyse the effects of the attention model, group-wise training approach, as well as the attention regularization described in Section~\ref{sec:robust} independently.
\begin{enumerate}
	\item ``Average pooling without attention (AP)'': We employ the average-pooling method as an important baseline here since it's commonly used for image classification on clean images without any noise-robust strategy.
	 The average pooling structure simply replaces the two 4096 dimensions fully-connected layers in VGG-16 model with an average pooling layer, followed by a softmax layer for classification.
	\item ``Random grouping training without attention (RGT)'': In this method, samples are randomly grouped during training, with the mean-pooling operation in Eq.~\ref{eq:sum_pool} to get the instance-level representation.
	\item ``Average pooling with attention (AP+AT)'':  Based on AP, the attention model is embedded in the network to test its ability to localize discriminative feature regions.
	\item ``Random grouping training with attention (RGT+AT)'': Attention is added to RGT.
	\item ``Average pooling with attention and regularizer (AP+AT+R)'': We further add the regularizer to AP+AT to evaluate its influence to cope with noisy labels.
	\item ``Random grouping training with attention and regularizer (RGT+AT+R)'': We test its performance on filtering out incorrectly labeled samples in each group as well as noisy local feature parts by adding the regularizer to RGT+AT.

\end{enumerate}

\subsection{Evaluation on the WebCars} \label{sec:cars}

We quantatively compare the methods described in Section~\ref{sec:implementation} and report the results in Table~\ref{tab:compare}. For RGT based methods, the group size is set to 2.

\begin{table}[htp]
	\centering
	\scalebox{1.0}
	{
		\begin{tabular}{l c}
			\hline
			methods&accuracy\\\hline
			AP &   66.86\%  \\
			RGT &  69.83\%   \\\hline
			AP+AT &  73.64\% \\
			RGT+AT & 76.58\%  \\\hline
			AP+AT+R & 70.77 \%\\
			RGT+AT+R & \textbf{78.44}\% \\\hline
		\end{tabular}
	}
	\caption{Comparison of classification results on the Compcars test set.}
	\label{tab:compare}
\end{table}

\noindent{\textbf{Average pooling vs. Random grouping training}}

By comparing the results of AP and RGT, we can see that the group-wise training can effectively suppress the influence of noise due to the improved labeling accuracy at the instance level. For this reason, the model can always learn some useful information from the correctly labeled samples in each group. In contrast, for training at the image level with no attention, the noisy labels will give networks misleading information that will harm the learning process.

\noindent{\textbf{Attention vs. without attention}}

For AP+AT and RGT+AT, the accuracy all improves by a large margin compared to AP and RGT respectively, which proves the effectiveness of the attention model employed. The attention model filters out uninformative parts of the feature maps for each sample and only let the useful parts flow through the latter network for classification. In this way, it works like a gate that can prevent the noisy regions of the feature representation from misleading the classifiers. A similar strategy is found effective on clean images for multi-label image classification~\cite{zhao2016regional}.

\noindent{\textbf{With vs. without regularizer}}

An interesting phenomenon we observe is that the accuracy for AP+AT drops significantly when using the noise regularizer, AP+AT+R. The reason is that the noise presents in both classes of object-of-interest and negative class, and consequently the image-level learning strategy confuses the network with how to classify the noise.  But this confusion doesn't exist in the group-level training approach, since very few training instances have incorrect labels after random grouping.  The reasons for adding noise regularizer is helpful for group-wise training are two-fold: First, the hinge loss regularizer forces the attention map not to concentrate on any feature regions of mislabeled samples, which results in a much cleaner group-level feature representation; Second, it helps the classifiers to distinguish the correctly labeled samples from the noise~\cite{girshick2014rich}. It's worth noticing that compared to utilizing clean images as constraint~\cite{xiao2015learning}, the negative samples are much easier to collect.

We consider two types of label noise defined in \cite{krause2015unreasonable}, which are called \emph{cross-domain} noise and \emph{cross-category} noise. The cross-domain noise is defined to be the portion of images that are not of any category in the fine-grained domain, \emph{i.e.} for cars, these images don't contain a car. In contrast, the cross-category noise is the mislabeled images within a fine-grained domain, \emph{i.e.} a car example with the wrong model label.

\begin{figure*}[tbp]
	\centering
	\resizebox{1.0\linewidth}{!}
	{\includegraphics{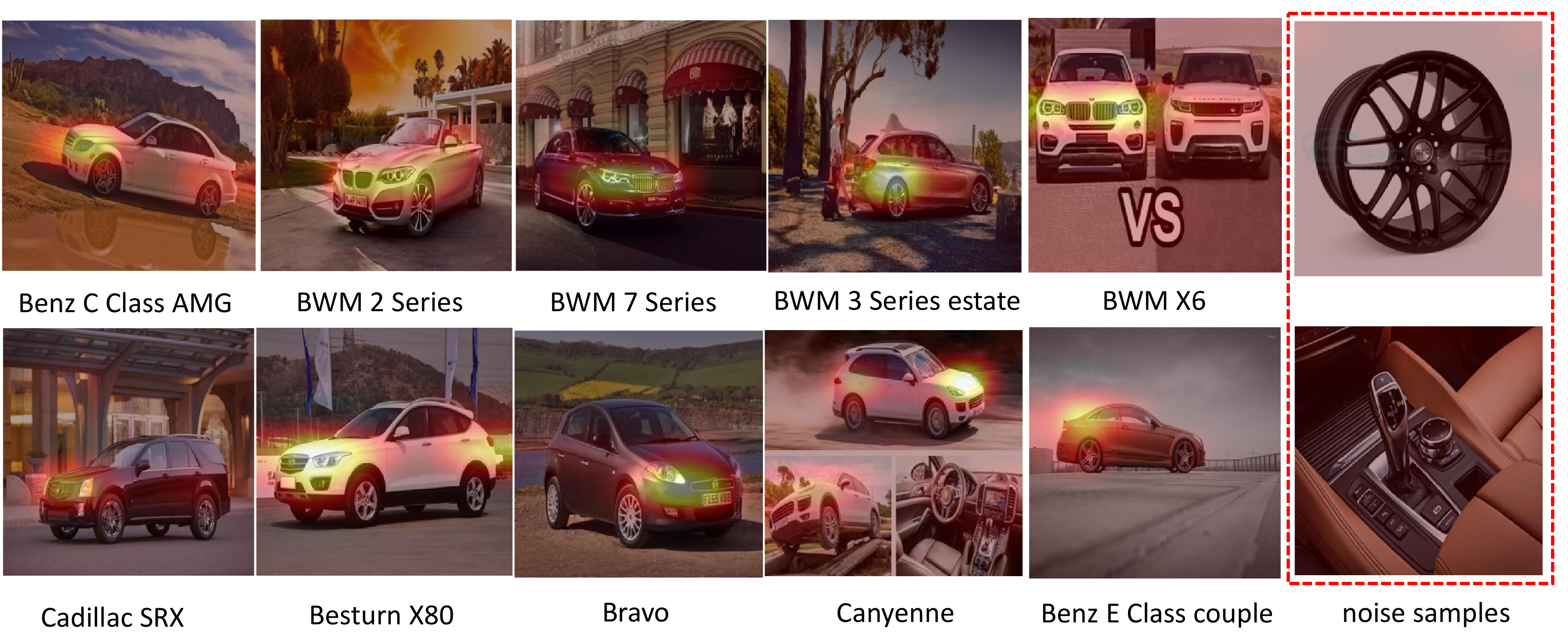} }
	\caption{Examples of the attention maps using the large-scale noisy fine-grained dataset described in Section~\ref{sec:datasets}. The brighter the region, the higher the attention scores. The examples in the red dotted box are mislabeled samples on the Web. }
	\label{fig:finegrained}
\end{figure*}

To better understand the proposed noise-robust mechanism, we provide qualitative visual results in Figure~\ref{fig:finegrained}. We see that the attention model mostly focuses on the discriminative parts in the front of or at the end of the cars. For some challenging examples, the correctly labeled car appears simultaneously with the cross-domain noise or cross-category noise in the same image. In this case, the attention model still successfully localizes to the correct parts. For the mislabeled samples, there's no object-of-interest to be concentrated on.

\subsection{Analysis of group size}\label{sec:group_size}

\begin{figure}[tbp]
	\centering
	\resizebox{\linewidth}{!}
	{\includegraphics[width=0.45\linewidth, height=0.3\linewidth]{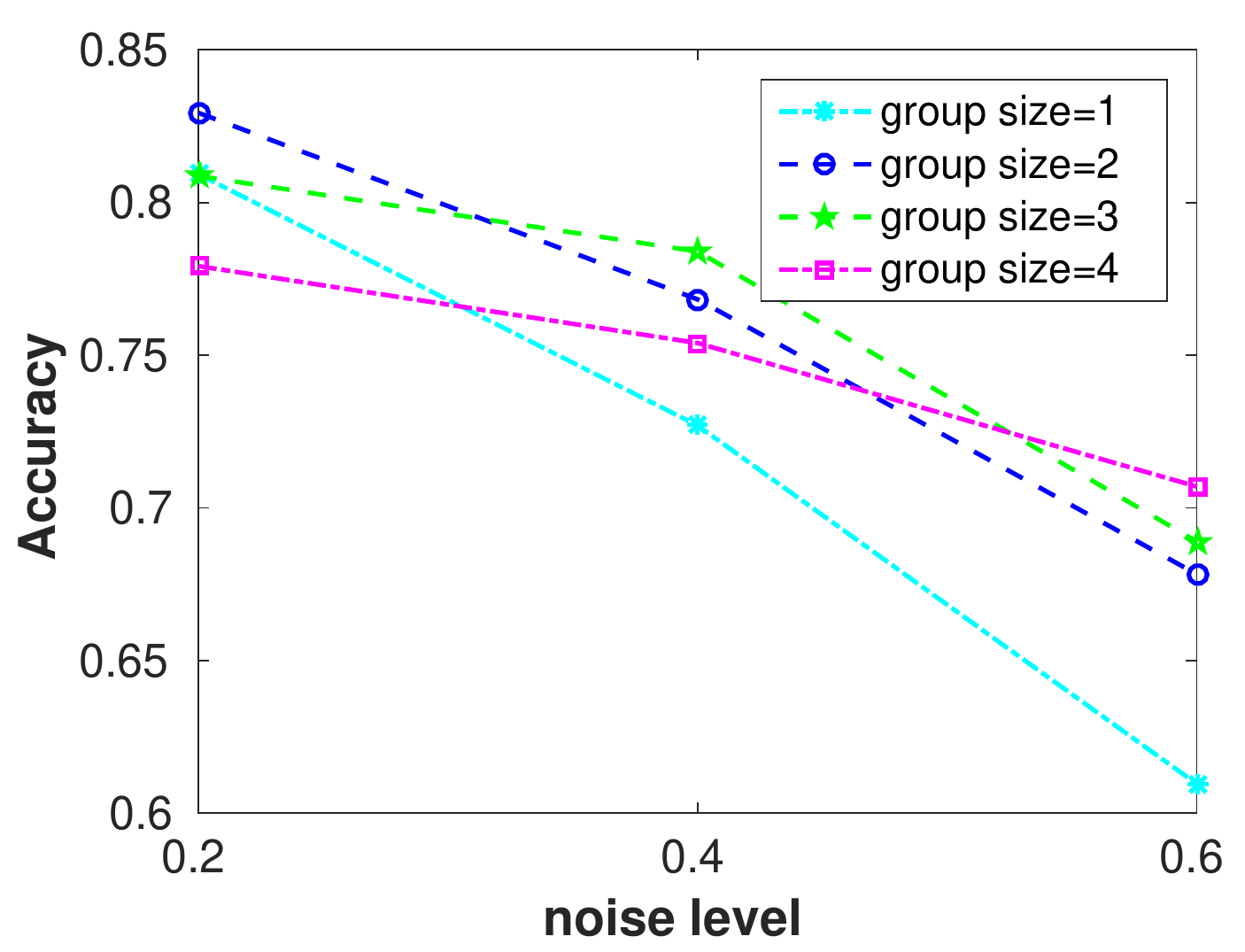}	}
	\caption{The classifcation accuracy under different group sizes of the proposed method.}
	\label{fig:group_size}
\end{figure}
In this section, we conduct a toy experiment to investigate the impact of the group size on our method (RGT+AT+R)\footnote{When group size equals 1, the method is equivalent to AP + AT + R. We empirically find adding the regularization term in this case will lead to inferior performance so we do not use the regularization term when group size equals 1.}. We randomly sample 100 car categories of the Compcars dataset and deliberately pollute the clean training data by adding cross-category noise and cross-domain noise in a proportion of 1:1. The total number of training images doesn't change. We then gradually increase the noise level from 0.2 to 0.6 and report the classification accuracy on the test set of Compcars using different group sizes. The results are shown in Figure~\ref{fig:group_size}. From Figure~\ref{fig:group_size}, we could make the following observations: (1) using group size $\ge$ 2 makes the network training more robust to noise. As can be seen, when the dataset contains a substantial amount noise label e.g. noise level = 0.6, the performance gap between group size = 1 and group size $\ge$ 2 can be larger than 10\%. (2) the optimal group size changes with the noise level. For example, when the noise level = 0.2, the optimal group size is 2 but when the noise level = 0.6, the optimal group size becomes 4. This observation could be partially explained by the analysis in section \ref{sec:weakly}, that is, the larger group size reduces the chance of having an incorrect label at the group-level. (3) Finally, we observe that larger $k$ does not always lead to better performance. As also mentioned in section \ref{sec:weakly}, we speculate that this is due to the fact that having a larger group will reduce the independency of grouped instance. For example, when having a larger $k$, the chance of two groups sharing one common image will grow significantly.

\begin{figure*}[tbp]
	\centering
	\resizebox{1.0\linewidth}{!}
	{\includegraphics{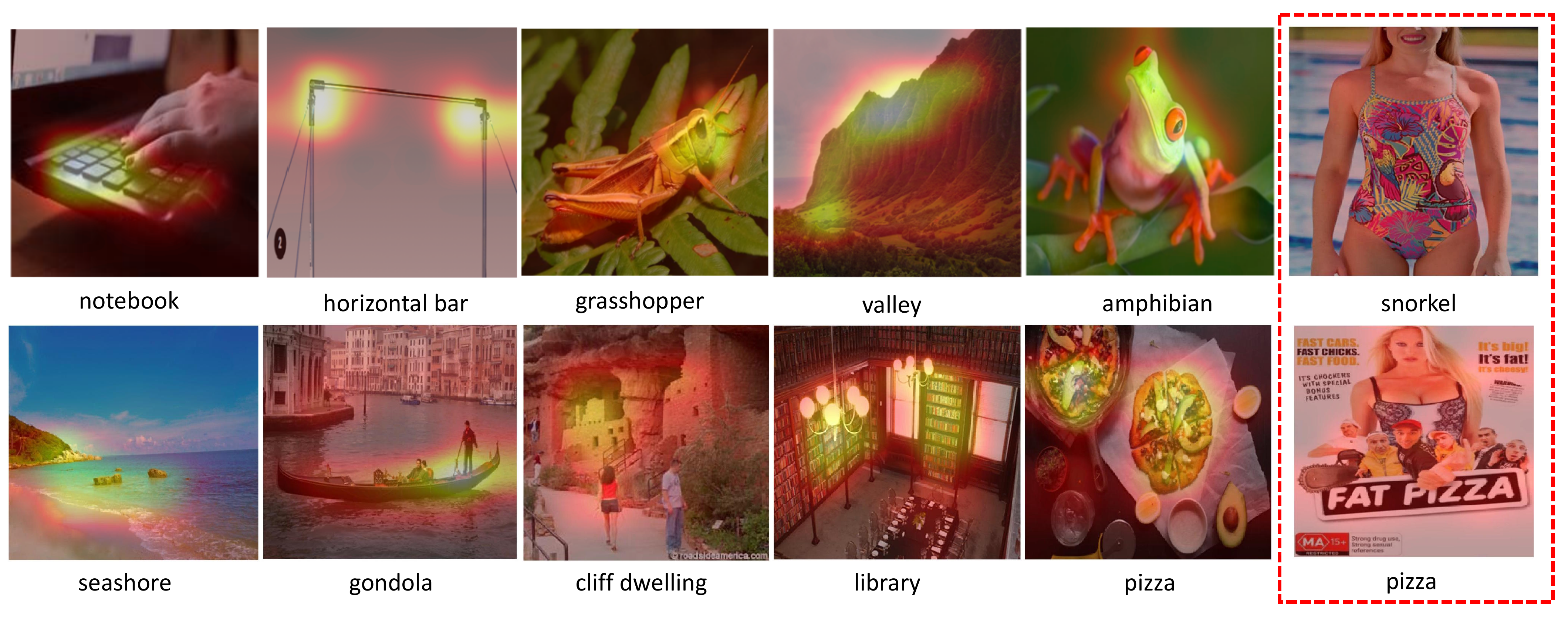} }
	\caption{Examples of where attention maps for the collected Web data with respect to ImageNet described in Section~\ref{sec:datasets}. The brighter the region, the higher the attention scores. The examples in the red dotted box are mislabeled samples on the Web.}
	\label{fig:imagenet}
\end{figure*}

\subsection{Web Images re-ranking} \label{sec:re-ranking}

To inspect whether the proposed method utilize the information from the correctly labeled data for training while ignoring the mislabeled ones, we now propose to re-rank the noisy training data used in Section~\ref{sec:group_size} according to their classification scores. The ideal case is that the highly ranked images are all correctly labeled ones while the low-ranking samples are mislabeled ones on the Web.
We compare three methods here, including AP, AP+AT as well as RGT+AT+R using different group sizes. The ground truth labels for correctly labeled images and mislabeled images are set to $+1$ and $-1$, respectively. Correctly labeled images are ranked high in the ground truth labels. Based on the learned models in Section~\ref{sec:group_size}, we first obtain the classification score for each training sample and rank the images in descending order based on their corresponding classification scores to get the predicted labels in each category. We then calculate the mean average precision (MAP) under different noise levels and group sizes. The mean average precision is obtained by averaging the precisions calculated at the total number of samples in different categories.

\begin{table}[htp]
	\centering
	\resizebox{1.0\linewidth}{!}
	{
		\begin{tabular}{l | c c c}
			\hline
			\backslashbox{methods}{noise level}&20 \%&40 \%&60 \%\\\hline
			AP & 93.72 &85.08 &74.42 \\\hline
			AP+AT & 96.71 &92.84 &90.56  \\\hline
			RGT+AT+R, group size=2 & \textbf{98.12} &95.81 &91.00 \\
			RGT+AT+R, group size=3 &97.71  &\textbf{95.93}  &91.04 \\
			RGT+AT+R, group size=4 & 97.95 &95.33 &\textbf{91.98}  \\\hline
		\end{tabular}
	}
	\caption{Comparison of mean average precisions \% using several methods under different noise levels.}
	\label{tab:re-rank}
\end{table}

From the table, we can see that for direct average pooling, the precision drops dramatically as the noise level increases. On the contrary, simply adding attention model only, the precision improves considerably especially when the noise level is high enough. For example, at the noise level 60 \%,  the precision gap is more than 15 \%. This result proves that selecting discriminative regions for each sample can effectively prevent noisy parts from impacting the final classification. By incorporating the group-wise training strategy, the performance further improves. This can be attributed to the highly accurate group-level labels used and the attention model for blocking the local features of mislabeled samples to generate the group-level representation. Overall, the proposed method is stable and performs well at different noise levels.

We also randomly select a car category and qualitatively evaluate the re-ranking performance at the noise level 0.4 (see Figure~\ref{fig:rerank}). The images are ranked in descending order based on their classification scores. We can see that only a pair of images are ranked incorrectly among the samples. From the results, we can expect that our method can further be used to assist collecting clean datasets or active learning.

\subsection{Evaluation on Web Images + ImageNet}  \label{sec:imagenet}
Apart from the challenging fine-grained classification task, the proposed method can also be generalized to a conventional classification task. We trained models from scratch using the noisy Web data with respect to ImageNet described in Section~\ref{sec:datasets} and test the performance on the ILSVRC2012 validation set.

\begin{table}[htp]
	\centering
	\scalebox{1.0}
	{
		\begin{tabular}{l c}
			\hline
			methods&accuracy\\\hline
			AP &  58.81\%  \\\hline
			AP+AT &  67.68\%   \\\hline
			RGT+AT+R, group size=2 &  \textbf{71.24}\% \\
			RGT+AT+R, group size=3 &  68.89\% \\
			RGT+AT+R, group size=4 &  66.23\% \\\hline
		\end{tabular}
	}
	\caption{Comparison of classification results on ILSVRC2012 test set.}
	\label{tab:imagenet}
\end{table}

From the results we can see that for the conventional image classification task with Web data, the proposed method still works much better than the directly average pooling baseline. By only applying the attention model on each sample to select discriminative feature regions for classification, the result improves by $\sim9\%$. By randomly generating groups online using reasonable group size and incorporating the regularizer, we get the best performance at the optimal group size 2, which confirms the conclusions in Section~\ref{sec:cars} and Section~\ref{sec:group_size}.

We also visualize some examples with their attention maps in Figure~\ref{fig:imagenet} using the best performed method RGT+AP+R with group size 2. The attention model attempts to localize the most discriminative parts for correctly labeled samples to push them far from the decision boundary. Samples in the red bounding box are mislabeled on the Web and the attention model finds no parts to concentrate on.

\section{Conclusion}
In this paper, we propose a weakly-supervised framework to learn visual representions from massive Web data without any human supervision. The proposed method can handle label noise effectively by two unified strategies. By randomly stacking training images into groups, the accuracy of the group-level labels improves. The attention model embedded further localizes the discriminative regions corresponding to correctly labeled samples across the combined feature maps for classification. The efficacy of our methods have been demonstrated by the extensive experiments.

\small{
	\bibliographystyle{ieee}
	\bibliography{draft}
}

\end{document}